\documentclass[sigconf]{acmart}
\copyrightyear{2023}
\acmYear{2023}
\setcopyright{acmlicensed}\acmConference[MM '23]{Proceedings of the 31st ACM International Conference on Multimedia}{October 29-November 3, 2023}{Ottawa, ON, Canada}
\acmBooktitle{Proceedings of the 31st ACM International Conference on Multimedia (MM '23), October 29-November 3, 2023, Ottawa, ON, Canada}
\acmPrice{15.00}
\acmDOI{10.1145/3581783.3612260}
\acmISBN{979-8-4007-0108-5/23/10}
\AtBeginDocument{%
  }

\usepackage{multirow}

\usepackage{colortbl}  
\usepackage{array}   
\definecolor{gray}{gray}{.9}
\definecolor{c1}{HTML}{EE9B01}
\definecolor{c2}{HTML}{2F70AF}
\definecolor{c3}{HTML}{BC0000}

\usepackage{pifont}
\newcommand{\cmark}{\ding{51}}%
\newcommand{\xmark}{\text{\ding{55}}}

\begin{document}

\title{UniNeXt: Exploring A Unified Architecture for Vision Recognition}


\author{Fangjian Lin}
\email{linfangjian01@163.com}
\orcid{0000-0001-9296-7531}
\affiliation{%
  \institution{Alibaba Group}
  \city{Hangzhou}
  \country{CN}
}

\author{Jianlong Yuan}
\authornote{Corresponding Author.}
\email{gongyuan.yjl@alibaba-inc.com}
\orcid{0000-0001-8306-2394}
\affiliation{%
  \institution{Alibaba Group}
  \city{Beijing}
  \country{CN}}

\author{Sitong Wang}
\email{stonewst@163.com}
\orcid{0000-0002-2830-2831}
\affiliation{%
  \institution{Alibaba Group}
  \city{Hangzhou}
  \country{CN}}

\author{Fan Wang}
\email{fan.w@alibaba-inc.com}
\orcid{0000-0001-7320-1119}
\affiliation{%
  \institution{Alibaba Group}
  \city{Hangzhou}
  \country{CN}}

\author{Zhibin Wang}
\email{zhibin.waz@alibaba-inc.com}
\orcid{0000-0001-7618-7973}
\affiliation{%
  \institution{Alibaba Group}
  \city{Hangzhou}
  \country{CN}
}

\begin{abstract}
Vision Transformers have shown great potential in computer vision tasks. 
Most recent works have focused on elaborating the spatial token mixer for performance gains. 
However, we observe that a well-designed general architecture can significantly improve the performance of the entire backbone, regardless of which spatial token mixer is equipped. 
In this paper, we propose UniNeXt, an improved general architecture for the vision backbone. 
To verify its effectiveness, we instantiate the spatial token mixer with various typical and modern designs, including both convolution and attention modules.
Compared with the architecture in which they are first proposed, our UniNeXt architecture can steadily boost the performance of all the spatial token mixers, and narrows the performance gap among them. 
Surprisingly, our UniNeXt equipped with naive local window attention even outperforms the previous state-of-the-art. 
Interestingly, the ranking of these spatial token mixers also changes under our UniNeXt, suggesting that an excellent spatial token mixer may be stifled due to a suboptimal general architecture, which further shows the importance of the study on the general architecture of vision backbone. 
Code is available at \href{https://github.com/jianlong-yuan/UniNeXt}{\textcolor[rgb]{0.88,0.0078,0.52}{UniNeXt}}.
\end{abstract}



\maketitle

\section{Introduction}
The architecture of Vision Transformer \cite{ViT} whose block contains a spatial token mixer and channel MLPs, has been proven promising. 
While subsequent works\cite{Swin, PVT, CSWin, focaltrans} have emphasized the importance of skillful spatial token mixer design in enhancing Transformer performance, the MetaFormer\cite{metaformer} argues that the overall double-jump connection architecture, which includes both spatial token and channel mixers, is the primary reason for the Transformer's success. This realization prompted a reconsideration of the Transformer architecture as a whole.

Most works have focused on elaborating the spatial token mixer for further improvements.
Some of them put their efforts into a well-designed attention mechanism by cross-window connection\cite{Swin, shuffle}, axial window\cite{CSWin, Pale, axwin}, dynamic window\cite{dgt, DAT}. 
In contrast, convolutional token mixers also gained much success via large kernel\cite{LargeKernel} and deformable kernel designs\cite{internimage}.
Moreover, some works have been dedicated to amalgamating the strengths of both convolution and attention by synthesizing them\cite{mobileformer,conformer}.

\begin{figure}[tp]
\centering
\includegraphics[width=1\linewidth]{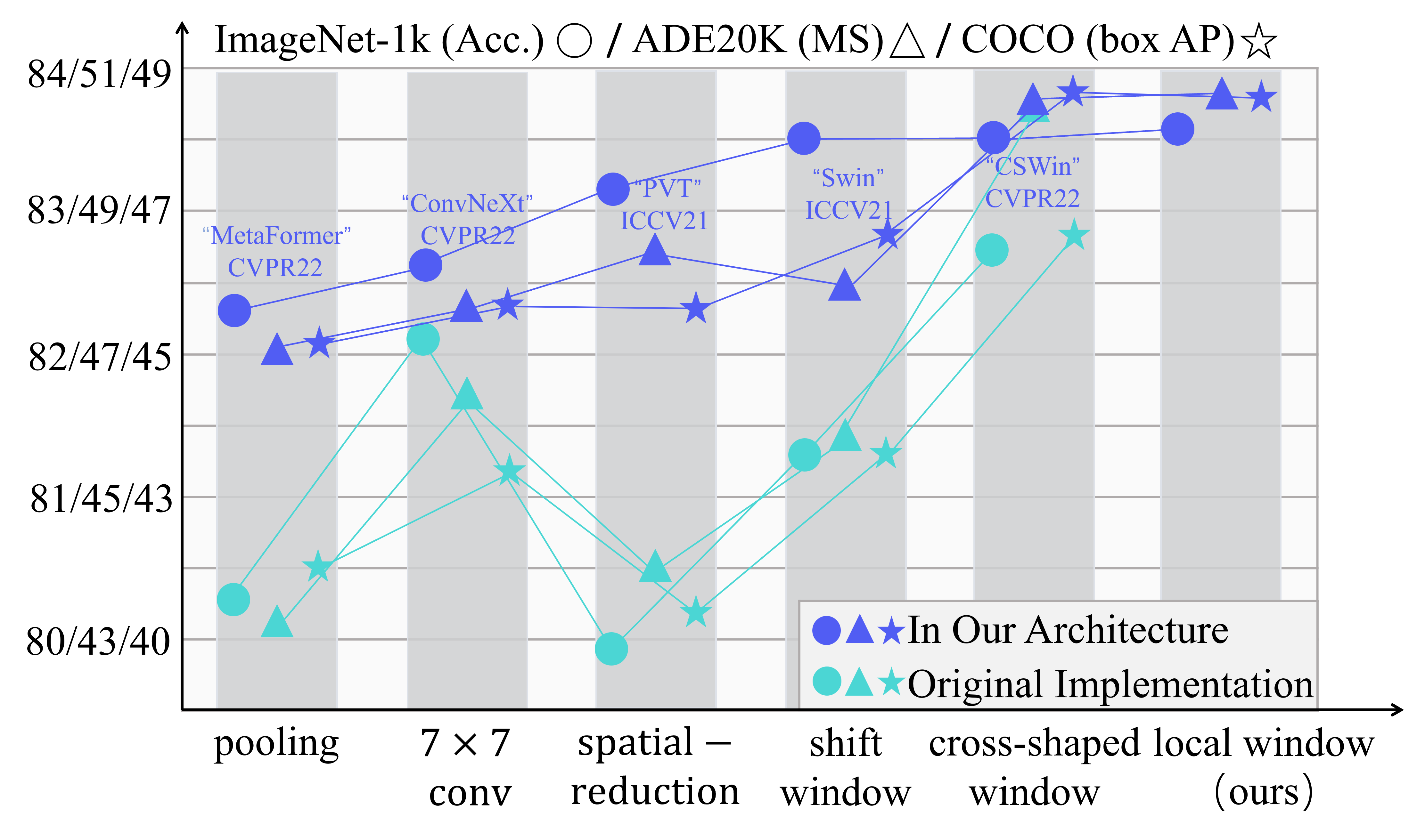} 
\caption{Comparison of different architectures. The results for the Tiny version are reported here. The Parameters and FLOPs are ignored here, see Table \ref{ablation_attn_mode} for more details. Different token mixers have received stable performance improvements with our architecture.}
\label{compare}
\end{figure}

While many researchers focus on designing novel spatial token mixers, we have discovered that the macro design of the entire network architecture is also crucial. As such, we propose a unified architecture that can be applied to different token mixers such as convolutional, attentional, and parameter-free ones. Our proposed architecture enables us to compare various mainstream spatial token mixers. As shown in Figure \ref{compare}, we get some surprising findings:
1) Our unified architecture can consistently improve performance across different spatial token mixers, surpassing the vanilla double-jump connection architecture.
2) Our proposed architecture reduces the gap in performance between the mainstream spatial token mixers.
3) The performance order of the different spatial token mixers changes under our unified architecture, such that even the most common window attention outperforms the shift window attention of Swin\cite{Swin}.

Our proposed unified architecture design hinges on our various jump connection modules. These modules include: High-dimensional Convolution (HdC), which enhances local modeling and extracts high-dimensional implicit features; Embedded Convolution (EC), which enhances the local modeling ability of spatial token mixers via a dual-branch rectification structure; and Post Convolution (PC), which performs additional local modeling on features emerging from each block. Additionally, we have designed a stem module that differs from previous work\cite{stem,cmt,dgt} to smooth the model features. 

We have evaluated our model on ImageNet-1k\cite{imagenet} classification, ADE20k\cite{ADE20K} semantic segmentation, and COCO\cite{coco} object detection and instance segmentation, and found that our unified architecture outperforms state-of-the-art results even using the simplest local window attention as the token mixer strategy. We have instantiated popular token mixer approaches\cite{metaformer,convnext,PVT,Swin,CSWin} within our unified framework, including parameter-free token mixer, convolution, local-attention, and global-attention, all of which resulted in stable improvements. Finally, we have investigated the incremental effect of additional convolution, measuring the overall effective receptive field (ERF)\cite{erf} of the model on ADE20k, and have found that the introduction of an additional 3$\times$3 convolution layer (the size of local window attention 11$\times$11) significantly enhances the effective receptive field.

\section{Related Work}
\subsection{General Architecture of Vision Backbone}
1) The VGG\cite{vgg}, while being a breakthrough in its time, ultimately failed to increase the depth of neural networks. The network architecture's key bottleneck was attributable to the large number of weights, making it computationally expensive and impractical for deeper networks. Despite this shortcoming, the VGG architecture laid the foundation for deeper networks by popularizing the idea of using small kernels in convolutional layers.

2) In contrast, the ResNet\cite{ResNet} supported deep networks by introducing a skip connection, which allowed network depth to increase without inducing vanishing gradients. By simplifying the learning task for deep models, the ResNet architecture achieved breakthrough results on the ImageNet dataset.

3) On the other hand, the Inception\cite{inception} opted to go wider, with a focus on multi-scale feature representation. This was achieved by utilizing convolutional layers with varying kernel sizes in parallel. The resulting network was computationally efficient and achieved state-of-the-art performance on ImageNet.

4) The Vision Transformer (ViT)\cite{ViT}, a recent development, has been proven to be a superior architecture by the MetaFormer\cite{metaformer}. This architecture eliminated the need for spatial convolutions by utilizing a sequence of tokens as input. The key innovation was putting the transformer architecture to work in computer vision tasks, and the ViT architecture has since been applied in a variety of vision tasks with excellent performance.

5) Dai at.el\cite{dtc} designed a unified architecture to provide a fair comparison for traditional and modern spatial token mixers.

Different from them, we aim to explore a superior general architecture for vision backbones.

\subsection{Spatial Token Mixer}
\noindent \textbf{Parameter-free Token Mixers.}
Recently, parameter-free token mixer strategies have been widely used in MLP-like architectures\cite{asmlp,cyclemlp}, which allow MLP architectures to fuse information from different locations through pixel shift operations, with their efficiency and parameter-free properties attracting researchers' attention.
MetaFormer\cite{metaformer} uses the pooling operation as a token mixer to accomplish the efficient design of Transformer architecture. And nowadays more novel techniques emerge that can bring further improvements to the parameter-free token mixer.

\noindent \textbf{Convolution-based Token Mixers.}
Convolution is massively used in vision tasks due to its sparse connectivity, weight sharing, and specific inductive bias properties. In recent years, the position of convolution in the vision backbone has been challenged by the vision Transformer. ConvNeXt\cite{convnext} uses a 7$\times$7 depth-wise convolution as the token mixer, which inherits and extends the Transformer architecture and lays the foundation for further development of the design of convolutional backbones. Not only the overall architecture changes but also some novel engineering designs can often bring significant improvements to the convolutional backbone.

\noindent \textbf{Attention-based Token Mixers.}
Since the Vision Transformer\cite{ViT} was proposed, the Attention-based Token mixer has received much attention due to its dynamic long-range spatial modeling capability. However, due to the quadratic complexity of attention, global attention tends to bring a large computational overhead. PVT\cite{PVT} reduces the computation of attention through the spatial-reduction mechanism. Swin Transformer\cite{Swin} restricts attention to local windows and enhances inter-window information interaction through shift transform, resulting in an efficiency and performance trade-off. To further expand the receptive field of window attention and improve the performance of the model in dense prediction tasks, CSWin\cite{CSWin} and Pale Transformer\cite{Pale} use bar window attention for spatial information aggregation.

\begin{figure}[!h]
\centering
\includegraphics[width=1.0\linewidth]{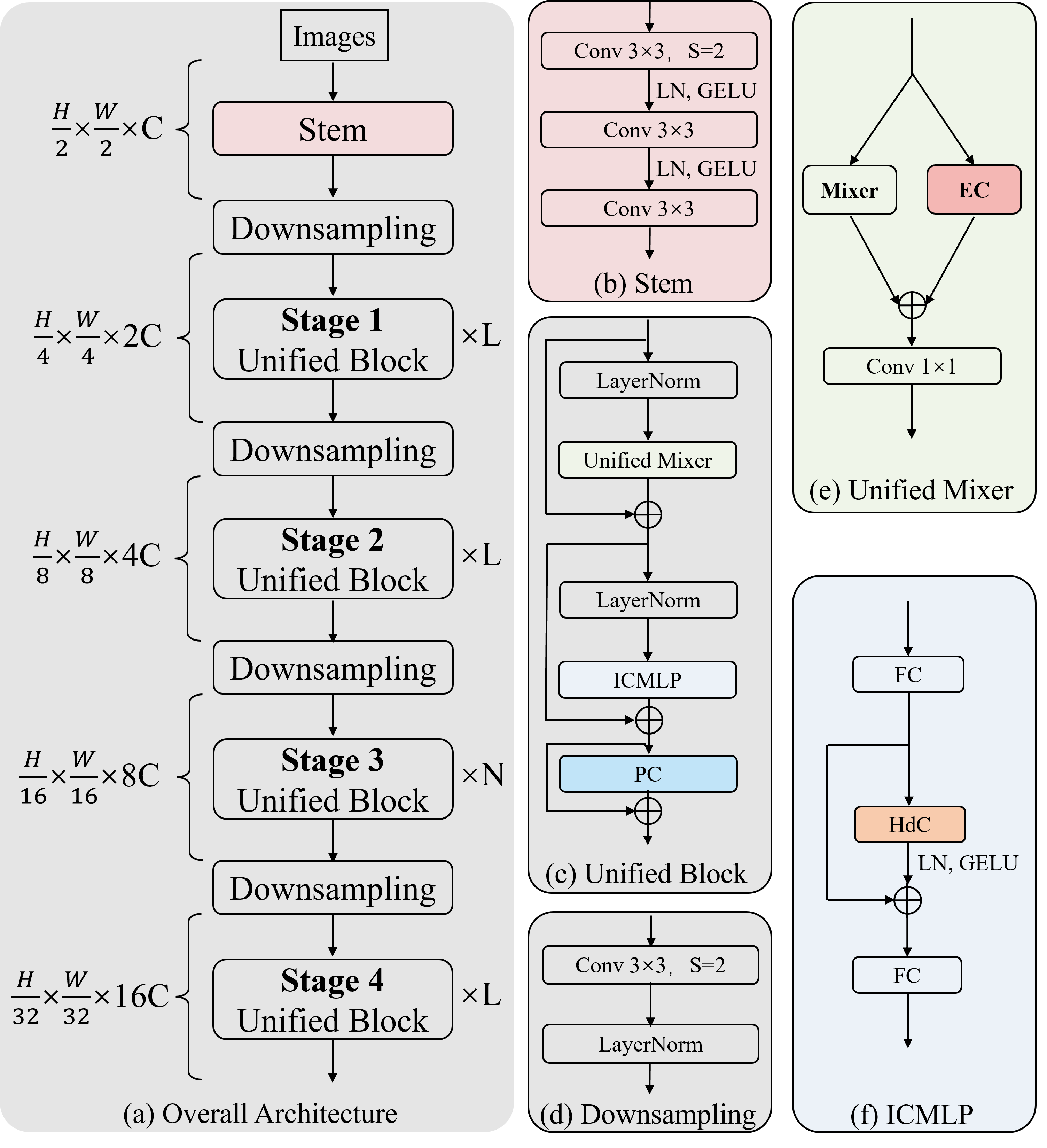} 
\caption{The overall framework of our UniNeXt. The Embedded Convolution (EC), High-dimensional Convolution (HdC), and Post Convolution (PC) are a 3$\times$3 depth-wise convolution. The ``Mixer'' includes parameter-free, convolution, and attention. The first FC layer in Inner Convolution MLP (ICMLP) performs channel dimension expansion.
}
\label{framework}
\end{figure}

\section{Method}
In this section, we first describe our overall architecture and variants. Then we describe in detail each of the proposed jump connection modules and spatial token mixer.

\subsection{Overall Architecture}
The main design idea of our unified architecture is to increase the inductive bias / locality by 3 means: a) adding a parallel EC branch to the spatial token mixer; b) adding an PC module behind the channel MLP; c) adding a 3$\times$3 dwconv to the FFN.

As shown in Figure \ref{framework}, similar to the classical work\cite{Swin,PVT,convnext}, UniNeXt is a pyramidal structure and consists of four hierarchical stages. 
Directly converting the image to the token for training tends to ignore the image structure information. The stem\cite{stem} can maintain the smoothness of structural features and perform progressive downsampling, so the image is first mapped through a stem layer to obtain a feature map with expanded channel dimension and reduced spatial resolution by twice.
Each stage contains a Downsampling layer and multiple Unified Blocks. In the Downsampling layer, the spatial downsampling ratio is 2 for each stage, and expand the channel dimension by twice. The output of the Downsampling layer is fed into multiple serially connected Unified blocks, where the number of tokens is kept consistent. Following \cite{Swin,CSWin}, finally we apply a global average pooling operation and a fully connected layer to perform the image classification task.

\subsection{Variants}

\begin{table}[tp]
\caption{Detailed configurations of different variants of our UniNeXt.}
    \resizebox{1.\columnwidth}{!}{
        \begin{tabular}{c|c|c|c|c}
            \toprule [0.15em]
            \begin{tabular}{c}Stage/Stride\end{tabular}
               & Layer
               & UniNeXt-T
               & UniNeXt-S
               & UniNeXt-B
            \\
            \hline
            Stride=2
               & Stem
               
               & $\begin{array}{c} \left[3\times3, 32, s\text{=}2\right] \times 1 \\ \left[3\times3, 32, s\text{=}1\right] \times 2 \end{array}$
               & $\begin{array}{c} \left[3\times3, 48, s\text{=}2\right] \times 1 \\ \left[3\times3, 48, s\text{=}1\right] \times 2 \end{array}$
               & $\begin{array}{c} \left[3\times3, 64, s\text{=}2\right] \times 1 \\ \left[3\times3, 64, s\text{=}1\right] \times 2 \end{array}$
            \\
            \hline

            \multirow{5}{*}{ \begin{tabular}{c} Stage 1 \\ Stride=4 \end{tabular}}
               & \begin{tabular}{c}Down-\\sampling\end{tabular}
               & $\begin{array}{c} \left[3\times3, 64, s\text{=}2\right] \times 1 \end{array}$
               & $\begin{array}{c} \left[3\times3, 96, s\text{=}2\right] \times 1 \end{array}$
               & $\begin{array}{c} \left[3\times3, 128, s\text{=}2\right] \times 1 \end{array}$
            \\
            \cline{2-5}
               & \begin{tabular}{c}Unified \\Block\end{tabular}
               & $\begin{bmatrix}\setlength{\arraycolsep}{1pt} \begin{array}{c}
                        H_1\text{=}2  \\
                        R_1\text{=}4 \\
                        W_1\text{=}7 \\
                        \widehat{W}_1\text{=}11  
                    \end{array} \end{bmatrix} \times 2$
               & $\begin{bmatrix}\setlength{\arraycolsep}{1pt} \begin{array}{c}
                        H_1\text{=}3  \\
                        R_1\text{=}4 \\
                        W_1\text{=}7 \\
                        \widehat{W}_1\text{=}11
                    \end{array} \end{bmatrix} \times 2$
               & $\begin{bmatrix}\setlength{\arraycolsep}{1pt} \begin{array}{c}
                        H_1\text{=}4  \\
                        R_1\text{=}4 \\
                        W_1\text{=}7 \\
                        \widehat{W}_1\text{=}11
                    \end{array} \end{bmatrix} \times 2$
            \\
            \hline
            \multirow{5}{*}{ \begin{tabular}{c} Stage 2 \\ Stride=8 \end{tabular}}
               & \begin{tabular}{c}Down-\\sampling\end{tabular}
               & $\begin{array}{c} \left[3\times3, 128, s\text{=}2\right] \times 1 \end{array}$
               & $\begin{array}{c} \left[3\times3, 192, s\text{=}2\right] \times 1 \end{array}$
               & $\begin{array}{c} \left[3\times3, 256, s\text{=}2\right] \times 1 \end{array}$
            \\
            \cline{2-5}
               & \begin{tabular}{c}Unified \\Block\end{tabular}
               & $\begin{bmatrix}\setlength{\arraycolsep}{1pt} \begin{array}{c}
                        H_2\text{=}4  \\
                        R_2\text{=}4 \\
                        W_2\text{=}7 \\
                        \widehat{W}_2\text{=}11
                    \end{array} \end{bmatrix} \times 2$
               & $\begin{bmatrix}\setlength{\arraycolsep}{1pt} \begin{array}{c}
                        H_2\text{=}6  \\
                        R_2\text{=}4 \\
                        W_2\text{=}7 \\
                        \widehat{W}_2\text{=}11
                    \end{array} \end{bmatrix} \times 2$
               & $\begin{bmatrix}\setlength{\arraycolsep}{1pt} \begin{array}{c}
                        H_2\text{=}8  \\
                        R_2\text{=}4 \\
                        W_2\text{=}7 \\
                        \widehat{W}_2\text{=}11
                    \end{array} \end{bmatrix} \times 2$
            \\
            \hline
            \multirow{5}{*}{ \begin{tabular}{c} Stage 3 \\ Stride\text{=}16 \end{tabular}}

               & \begin{tabular}{c}Down-\\sampling\end{tabular}
               & $\begin{array}{c} \left[3\times3, 256, s\text{=}2\right] \times 1 \end{array}$
               & $\begin{array}{c} \left[3\times3, 384, s\text{=}2\right] \times 1 \end{array}$
               & $\begin{array}{c} \left[3\times3, 512, s\text{=}2\right] \times 1 \end{array}$
            \\

            \cline{2-5}
               & \begin{tabular}{c}Unified \\Block\end{tabular}
               & $\begin{bmatrix}\setlength{\arraycolsep}{1pt} \begin{array}{c}
                        H_3\text{=}8  \\
                        R_3\text{=}4 \\
                        W_3\text{=}7 \\
                        \widehat{W}_3\text{=}11
                    \end{array} \end{bmatrix} \times 18$
               & $\begin{bmatrix}\setlength{\arraycolsep}{1pt} \begin{array}{c}
                        H_3\text{=}12  \\
                        R_3\text{=}4 \\
                        W_3\text{=}7 \\
                        \widehat{W}_3\text{=}11
                    \end{array} \end{bmatrix} \times 18$
               & $\begin{bmatrix}\setlength{\arraycolsep}{1pt} \begin{array}{c}
                        H_3\text{=}16  \\
                        R_3\text{=}4 \\
                        W_3\text{=}7 \\
                        \widehat{W}_3\text{=}11
                    \end{array} \end{bmatrix} \times 18$
            \\

            \hline
            \multirow{5}{*}{ \begin{tabular}{c} Stage 4 \\ Stride\text{=}32 \end{tabular}}

               & \begin{tabular}{c}Down-\\sampling\end{tabular}
               & $\begin{array}{c} \left[3\times3, 512, s\text{=}2\right] \times 1 \end{array}$
               & $\begin{array}{c} \left[3\times3, 768, s\text{=}2\right] \times 1 \end{array}$
               & $\begin{array}{c} \left[3\times3, 1024, s\text{=}2\right] \times 1 \end{array}$
            \\

            \cline{2-5}
               & \begin{tabular}{c}Unified \\Block\end{tabular}
               & $\begin{bmatrix}\setlength{\arraycolsep}{1pt} \begin{array}{c}
                        H_4\text{=}16  \\
                        R_4\text{=}4 \\
                        W_4\text{=}global \\
                        \widehat{W}_4\text{=}global
                    \end{array} \end{bmatrix} \times 2$
               & $\begin{bmatrix}\setlength{\arraycolsep}{1pt} \begin{array}{c}
                        H_4\text{=}24  \\
                        R_4\text{=}4 \\
                        W_4\text{=}global \\
                        \widehat{W}_4\text{=}global
                    \end{array} \end{bmatrix} \times 2$
               & $\begin{bmatrix}\setlength{\arraycolsep}{1pt} \begin{array}{c}
                        H_4\text{=}32  \\
                        R_4\text{=}4 \\
                        W_4\text{=}global \\
                        \widehat{W}_4\text{=}global
                    \end{array} \end{bmatrix} \times 2$
            \\
            \bottomrule[0.15em]
        \end{tabular}
    }
    \label{tab:arch}
\end{table}
As shown in Table \ref{tab:arch}, we scaled the UniNeXt to obtain three variants of different sizes, including UniNeXt-T, UniNeXt-S, and UniNeXt-B. The hyper-parameters in the i-th stage of the table are represented as follows:
\begin{itemize}
\item $H_i$: the head number for the local window attention,
\item $R_i$: the expansion ratio for the ICMLP module,
\item $W_i$: the window size for the local window attention in the image classification task,
\item $\widehat{W}_i$: the window size for the local window attention in the dense prediction tasks, including segmentation and detection.
\end{itemize}
The $\left[3\times3, 32, s\text{=}2\right] \times 1$ in Table \ref{tab:arch} represents a 3$\times$3 convolution with 32 output channels and stride\text{=}2.
Note that all variants have the same depth, expansion ratio, and window size, the difference lies in the channel dimension and head number.

\subsection{UniNeXt Block}

\noindent \textbf{High-dimensional Convolution (HdC).}
Researchers\cite{segformer,cmt} have found that adding a lightweight 3$\times$3 depth-wise convolution to the MLP can improve performance. We inherit and extend this convolution embedding mechanism. First, the feature dimensions are mapped using the first linear layer of the MLP to obtain high-dimensional feature $\mathcal{F}$. Then a 3$\times$3 depth-wise convolution is used for spatial local fusion to encode high-dimensional implicit features, which can bring efficiency gains. The forward pass can be formulated as follows:
\begin{equation} 
    \begin{aligned}
        \mathcal{F}^{'} = \mathcal{F} + GELU(LN(\phi(dwconv(\widetilde{\phi}(\mathcal{F}))))).
    \end{aligned}
    \label{eq:hdc}
\end{equation}
Where $\phi$ denotes the flattened operation upon the spatial dimension, and $\widetilde{\phi}$ means reverse operation.

\noindent \textbf{Embedded Convolution (EC).} 
The motivation behind designing Embedded Convolution (EC) has two main aspects. Firstly, it enhances the model's inductive bias, which is crucial for learning and generalization. Secondly, EC is compatible with all token-mixers, providing flexibility and ease of implementation across various architectures.

As shown in Figure \ref{framework}, for the input token mixer the feature map $\mathcal{F}$, the formula  can be described as:
\begin{equation} 
    \begin{aligned}
        \mathcal{F}^{'} =  token\text{-}mixer(\mathcal{F}) + dwconv(\mathcal{F}).
    \end{aligned}
    \label{eq:mixer1}
\end{equation}
where the token mixer includes parameter-free and convolution and the dwconv represents a lightweight 3$\times$3 depth-wise convolution. For attention, we perform depth-wise convolution directly on the value instead of after the window transformation, making it universally applicable to all kinds of attention. For input into attention's feature map $\mathcal{F}$, the forward pass can be described as:
\begin{equation} 
    \begin{aligned}
        Q, K, V = \Phi_q(\mathcal{F}), \Phi_k(\mathcal{F}), \Phi_v(\mathcal{F}),
    \end{aligned}
    \label{eq:qkv}
\end{equation}
\begin{equation} 
    \begin{aligned}
        \mathcal{F}^{'} = Attention(Q, K, V) + \phi(dwconv(\widetilde{\phi}(V))).
    \end{aligned}
\end{equation}
Where $\Phi$ represents a fully connected layer, $\phi$ and $\widetilde{\phi}$ can be found in Equation \ref{eq:hdc}. 
We have demonstrated through extensive ablation studies that Embedded Convolution (EC) can bring performance improvements to the model.

\noindent \textbf{Post Convolution (PC).}
To further enhance the convolution embedding bias and enhance the local representation, we propose the Post Convolution (PC), which can greatly improve the model performance, especially in dense prediction tasks, by introducing a lightweight 3$\times$3 depth-wise convolution. Specifically, we first convert the token $\mathcal{F}\in \mathbb{R}^{B \times N \times C}$ to a 2-D image representation, then perform depth-wise convolution for local context fusion, then perform the flattening operation, finally use a residual connection to prevent over-scaling of weights:
\begin{equation} 
    \begin{aligned}
        \mathcal{F}^{'} = \mathcal{F} + \phi(dwconv(\widetilde{\phi}(\mathcal{F}))).
    \end{aligned}
\end{equation}

\noindent \textbf{Stem.}
Unlike previous stem\cite{cmt,dgt,stem} architectures that use BatchNorm\cite{bn} (BN) to perform normalization, we are inspired by ConvNeXt\cite{convnext} and Vision Transformer\cite{ViT} and use Layer Normalization\cite{ln}(LN) instead of BN to perform normalization so that the normalization operation is consistent throughout the model, which helps the model train better and produce better performance. The detailed architecture can be seen in Figure \ref{framework}.

\begin{table*}[tp]
\caption{\label{tab:det}Object detection and instance segmentation performance on the COCO val2017 with the Mask R-CNN framework. The FLOPs (G) are measured at resolution $800\times 1280$, and the models are pre-trained on the ImageNet-1K.}
\renewcommand\arraystretch{1}
\begin{center}
\setlength{\tabcolsep}{1mm}{
\begin{tabular}{l@{\hspace{3.2pt}}|c@{\hspace{3.2pt}}|c|c|c|c|c|c|c}
\toprule[0.15em]
\multirow{2}{*}{Backbone} & \multirow{2}{*}{\#Params} & \multirow{2}{*}{\#FLOPs} & \multicolumn{6}{c}{Mask R-CNN (1x)} \\
 & & & $AP^b$ & $AP^b_{50}$ & $AP^b_{75}$ & $AP^m$ & $AP^m_{50}$ & $AP^m_{75}$\\
\midrule
PVT-S\cite{PVT}           & 44M & 245G  & 40.4 & 62.9 & 43.8 & 37.8 & 60.1 & 40.3 \\
Swin-T\cite{Swin}         & 48M & 264G  & 43.7 & 66.6 & 47.6 & 39.8 & 63.3 & 42.7\\
ConvNeXt-T\cite{convnext} & 48M & 262G  & 44.2 & 66.6 & 48.3 & 40.1 & 63.3 & 42.8 \\
CSWin-T\cite{CSWin}       & 42M & 279G  & 46.7 & 68.6 & 51.3 & 42.2 & 65.6 & 45.4 \\
\rowcolor{gray}
$\textbf{UniNeXt-T}^{\dag}$ (ours)          & 43M & 266G  & 48.6 & 70.6 & 53.4 & 43.4 & 67.6 & 46.7 \\
\rowcolor{gray}
$\textbf{UniNeXt-T}^{*}$ (ours)          & 43M & 282G  & 48.7 & 71.2 & 53.3 & 43.6 & 67.7 & 46.5 \\
\midrule
PVT-M\cite{PVT}            & 64M & 302G  & 42.0 & 64.4 & 45.6 & 39.0 & 61.6 & 42.1\\
Swin-S\cite{Swin}          & 69M & 354G  & 44.8 & 66.6 & 48.9 & 40.9 & 63.4 & 44.2 \\
ConvNeXt-S\cite{convnext}  & 70M & 348G  & 45.4 & 67.9 & 50.0 & 41.8 & 65.2 & 45.1\\
CSWin-S\cite{CSWin}        & 54M & 342G  & 47.9 & 70.1 & 52.6 & 43.2 & 67.1 & 46.2\\
\rowcolor{gray}
$\textbf{UniNeXt-S}^{\dag}$ (ours)           & 71M & 333G  & 49.0 & 71.3 & 54.1 & 43.7 & 68.0 & 46.9 \\
\rowcolor{gray}
$\textbf{UniNeXt-S}^{*}$ (ours)          & 71M & 353G  & 49.2 & 71.4 & 54.3 & 43.8 & 68.3 & 47.0 \\
\midrule
PVT-L\cite{PVT}            & 81M  & 364G  & 42.9 & 65.0 & 46.6 & 39.5 & 61.9 & 42.5 \\
Swin-B\cite{Swin}          & 107M & 496G  & 46.9 & ---- & ---- & 42.3 & ---- & ---- \\
ConvNeXt-B\cite{convnext}  & 108M & 486G  & 47.0 & 69.4 & 51.7 & 42.7 & 66.3 & 46.0  \\
CSWin-B\cite{CSWin}        & 97M  & 526G  & 48.7 & 70.4 & 53.9 & 43.9 & 67.8 & 47.3 \\
\rowcolor{gray}
$\textbf{UniNeXt-B}^{\dag}$ (ours)           & 111M & 460G  & 49.3 & 71.4 & 54.1 & 43.9 & 68.3 & 47.3 \\
\rowcolor{gray}
$\textbf{UniNeXt-B}^{*}$ (ours)          & 111M & 489G  & 49.8 & 72.0 & 54.8 & 44.5 & 68.9 & 48.0 \\
\bottomrule[0.15em]
\end{tabular}
}
\end{center}
\end{table*}

\noindent \textbf{Unified Mixer.}
\begin{figure}[!h]
\centering
\includegraphics[width=1\linewidth]{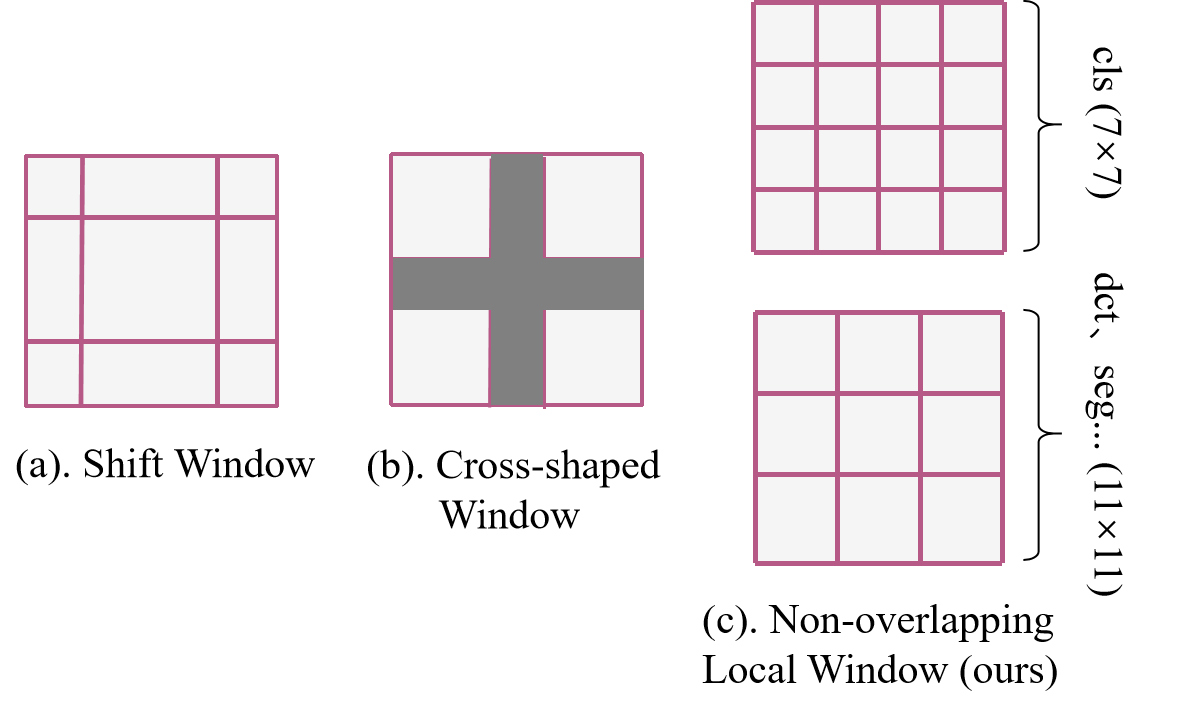} 
\caption{Illustration of different self-attention mechanisms. In our non-overlapping local window, the window size\text{=}7$\times$7 for image classification and window size\text{=}11$\times$11 for the dense prediction task. Note that the last stage of the model uses global attention.}
\label{div}
\end{figure}
By instantiating different token mixers (e.g., parameter-free, convolution and attention) in our architecture, the model performance steadily improves. In addition, using just the plain local window attention in our framework, outperform previous sota results. As shown in Figure \ref{div}, we first perform non-overlapping window partition on the feature map, using window size\text{=}7 for image classification and window size\text{=}11 for downstream tasks (e.g., detection and segmentation), more detailed ablation studies can be found in Table \ref{ablation_window_size}. Then perform multi-head self-attention:
\begin{equation} 
    \begin{aligned}
        \mathcal{F}^{'} = \text{SoftMax}(\frac{ Q K^T }{\sqrt D}) \cdot V.
    \end{aligned}
    \label{eq:attn}
\end{equation}
where Q, K, and V are generated as in Eq. \ref{eq:qkv}. Note that unlike ViT\cite{ViT} and Swin\cite{Swin}, we do not use additional absolute position encoding or relative position encoding, since the proposed convolution embedding technique is sufficient to represent the positional relationships between tokens.

\section{Experiments}
We first compare our UniNeXt with the state-of-the-art backbones on ImageNet-1K\cite{imagenet} for image classification. To further demonstrate the effectiveness and generalization of our backbone, we conduct experiments on ADE20K\cite{ADE20K} for semantic segmentation and COCO\cite{coco} for object detection \& instance segmentation. Then, we instantiated the classical spatial token mixer\cite{metaformer,convnext,PVT,Swin,CSWin} on our architecture and conducted some experiments, all of which resulted in greater improvements compared to the original architecture. Finally, we dig into the design of key components of our UniNeXt to better understand the method.
\subsection{Image Classification on ImageNet-1K}
\noindent \textbf{Settings.} 
All the variants are trained from scratch on 32 A10 GPUs with a total batch size of 2048. We employ an AdamW\cite{AdamW} optimizer for 300 epochs using a cosine decay learning rate scheduler and 10 epochs of linear warm-up. An initial learning rate of 0.002 is used. A weight decay of 0.05 for Tiny and Small, 0.1 for Base. We follow the data augmentation of Swin\cite{Swin} and CSWin\cite{CSWin} and use the EMA\cite{EMA} update strategy. Both the training and evaluation are conducted with the input size of $224\times224$ on the ImageNet-1K dataset. 

\noindent \textbf{Results.}
Table \ref{EXP_CLS} compares the performance of our UniNeXt with the state-of-the-art backbones on the ImageNet-1K validation set. 
\begin{table}[!h]
\caption{\label{EXP_CLS}Image classification performance on the ImageNet-1k validation set, where $\dag$ represents the use of local window attention and $*$ represents cross-shaped window attention.
} 
\newcommand{\tabincell}[2]{\begin{tabular}{@{}#1@{}}#2\end{tabular}}
\centering
\resizebox{0.49\textwidth}{!}{
    \begin{tabular}{l|cc|c}
    \toprule[0.15em]
\multirow{2}{*}{Backbone} & \multirow{2}{*}{\#Params} & \multirow{2}{*}{\#FLOPs} & \multirow{2}{*}{\tabincell{l}{\ \ Top-1 \\Acc. (\%)}} \\ 
    &      &     & \\
    \midrule[0.5pt] 
    PVT-S \cite{PVT}                 & 25M      & 3.8G   & 79.8         \\
    PoolFormer-S24 \cite{metaformer} & 21M      & 3.5G   & 80.3         \\
    Swin-T \cite{Swin}               & 29M      & 4.5G   & 81.3         \\
    Twins-SVT-S \cite{Twins}         & 24M      & 2.8G   & 81.3          \\
    PVTv2-B2 \cite{PVTv2}            & 25M      & 4.0G   & 82.0         \\
    ConvNeXt-T \cite{convnext}       & 29M      & 4.5G   & 82.1         \\
    Focal-T \cite{Focal}             & 29M      & 4.9G   & 82.2         \\
    Shuffle-T \cite{shuffle}         & 29M      & 4.6G   & 82.5         \\
    CSWin-T \cite{CSWin}             & 23M      & 4.3G   & 82.7         \\
    \rowcolor{gray}
    $\textbf{UniNeXt-T}^{*}$ (ours)    & 24M      & 4.3G   & \textbf{83.5}\\ 
    \rowcolor{gray}
    $\textbf{UniNeXt-T}^{\dag}$ (ours)        & 24M      & 4.3G   & \textbf{83.6}\\ 
    \midrule[0.5pt]
    PVT-M \cite{PVT}                 & 44M       & 6.7G  & 81.2         \\
    PoolFormer-M36 \cite{metaformer} & 56M       & 9.0G  & 82.1         \\
    Swin-S \cite{Swin}               & 50M       & 8.7G  & 83.0         \\
    Twins-SVT-B \cite{Twins}         & 56M       & 8.3G  & 83.1         \\
    ConvNeXt-S \cite{convnext}       & 50M       & 8.7G  & 83.1         \\
    PVTv2-B3 \cite{PVTv2}            & 45M       & 6.9G  & 83.2         \\
    Focal-S \cite{Focal}             & 51M       & 9.1G  & 83.5         \\
    Shuffle-S \cite{shuffle}         & 50M       & 8.9G  & 83.5         \\
    CSWin-S \cite{CSWin}             & 35M       & 6.9G  & 83.6         \\
    \rowcolor{gray}
    $\textbf{UniNeXt-S}^{\dag}$ (ours)        & 51M       & 9.5G  & 84.1        \\
    \rowcolor{gray}
    $\textbf{UniNeXt-S}^{*}$ (ours)  & 51M       & 9.6G  & 84.3        \\
    \midrule[0.5pt]
    PVT-L \cite{PVT}                 & 61M       & 9.8G    & 81.7       \\
    PoolFormer-M48 \cite{metaformer} & 73M       & 11.8G   & 82.5       \\
    Swin-B \cite{Swin}               & 88M       & 15.4G   & 83.3       \\
    Twins-SVT-L \cite{Twins}         & 99M       & 14.8G   & 83.3       \\
    PVTv2-B5 \cite{PVTv2}            & 82M       & 11.8G   & 83.8       \\
    Focal-B \cite{Focal}             & 90M       & 16.0G   & 83.8       \\
    ConvNeXt-B \cite{convnext}       & 89M       & 15.4G   & 83.8       \\
    Shuffle-B \cite{shuffle}         & 88M       & 15.6G   & 84.0       \\
    CSWin-B \cite{CSWin}             & 78M       & 15.0G   & 84.2       \\
    \rowcolor{gray}
    $\textbf{UniNeXt-B}^{\dag}$ (ours)        & 91M       & 17.1G   & 84.4       \\
    \rowcolor{gray}
    $\textbf{UniNeXt-B}^{*}$ (ours)  & 91M       & 17.3G  & 84.7        \\
    \bottomrule[0.15em]
\end{tabular}}
\end{table}

Compared with modern CNNs, our UniNeXt variants with local window attention are +1.5\%, +1.0\%, and +0.6\% better than the well-known ConvNeXt \cite{convnext}, respectively, under similar computation complexity. 
Meanwhile, our UniNeXt with local window attention outperforms the state-of-the-art Transformer-based backbones, which are +0.9\%, +0.5\%, and +0.2\% higher than CSWin Transformer\cite{CSWin}. In addition, our UniNeXt using plain local window attention is +2.3\%, +1.1\%, and +1.1\% better than Swin Transformer\cite{Swin} with shift window attention under the same window size\text{=}7$\times$7.
Finally, when UniNeXt is equipped with cross-shaped window attention, the base model achieves 84.7\% Top-1 Acc.

\begin{table}[!h]
\caption{\label{EXP_ade20k}Comparisons of different backbones with UperNet as decoder on ADE20K for semantic segmentation. All backbones are pretrained on ImageNet-1K with the size of $224\times224$. FLOPs are calculated with a resolution of $512\times2048$. 
}
\newcommand{\tabincell}[2]{\begin{tabular}{@{}#1@{}}#2\end{tabular}}
\centering   
\resizebox{0.49\textwidth}{!}{
    \begin{tabular}{l|c|cc|ccc}
    \toprule[0.15em]
    \multirow{2.2}{*}{\tabincell{l}{Backbone}} &\multirow{2}{*}{\tabincell{c}{crop \\ size}} & \multirow{2.2}{*}{\tabincell{l}{\#Params}}   &\multirow{2.2}{*}{\tabincell{l}{\#FLOPs}}  &\multirow{2}{*}{\tabincell{c}{mIoU \\SS}} &\multirow{2}{*}{\tabincell{c}{mIoU \\ MS}}         \\
     ~             & ~   & ~  & ~      & ~   \\
    \midrule[0.5pt]
    Swin-T \cite{Swin}              &$512^{2}$  & 60M  & 945G  & 44.5 & 45.8 \\
    ConvNeXt-T\cite{convnext}       &$512^{2}$  & 60M  & 939G  & 46.0 & 46.7 \\
    CSWin-T \cite{CSWin}            &$512^{2}$  & 60M  & 959G  & 49.3 & 50.4 \\
    \rowcolor{gray}
    $\textbf{UniNeXt-T}^{*}$(ours)                 &$512^{2}$  & 53M  & 961G   & 49.9 & 50.5 \\
    \rowcolor{gray}
    $\textbf{UniNeXt-T}^{\dag}$(ours)                 &$512^{2}$  & 53M  & 942G   & 49.7 & 50.6 \\
    \toprule[0.5pt]
    Swin-S \cite{Swin}              &$512^{2}$  & 81M  & 1038G & 47.6 & 49.5 \\
    ConvNeXt-S\cite{convnext}       &$512^{2}$  & 82M  & 1027G & 48.7 & 49.6 \\
    CSWin-S \cite{CSWin}            &$512^{2}$  & 65M  & 1027G & 50.0 & 50.8 \\
    \rowcolor{gray}
    $\textbf{UniNeXt-S}^{\dag}$(ours)                 &$512^{2}$  & 83M  & 1010G  & 51.0 & 51.8 \\
    \rowcolor{gray}
    $\textbf{UniNeXt-S}^{*}$(ours)                 &$512^{2}$  & 83M  & 1040G   & 51.5 & 52.4 \\
    \toprule[0.5pt]
    Swin-B \cite{Swin}              &$512^{2}$  & 121M & 1188G & 48.1	& 49.7 \\
    ConvNeXt-B\cite{convnext}       &$512^{2}$  & 122M & 1170G & 49.1 & 49.9 \\
    CSWin-B \cite{CSWin}            &$512^{2}$  & 109M & 1222G & 50.8 & 51.7 \\
    \rowcolor{gray}
    $\textbf{UniNeXt-B}^{\dag}$(ours)                 &$512^{2}$  & 124M & 1142G  & 51.4 & 52.2 \\
    \rowcolor{gray}
    $\textbf{UniNeXt-B}^{*}$(ours)                 &$512^{2}$  & 124M  & 1175G   & 51.6 & 52.5 \\
    \bottomrule[0.15em]
    \end{tabular}}
\end{table}

\begin{table*}[tp]
\caption{\label{ablation_attn_mode} Comparison with different spatial token mixer. $\color{c1}{\bullet}$, $\color{c2}{\bullet}$ and $\color{c3}{\bullet}$ represent the token mixer with parameter-free, convolution, and attention, respectively. Where the image sizes for training and inference are the same as in the experimental settings.}
\centering
\resizebox{1\linewidth}{!}{
\begin{tabular}{l|c|ccc|cccc|cccc}
\toprule[0.15em]
\multirow{2.2}{*}{Framework} &\multirow{2.2}{*}{Token Mixer} & \multicolumn{3}{c}{ImageNet-1K} & \multicolumn{4}{c}{ADE20K} & \multicolumn{4}{c}{COCO} \\
            & &\#Params &\#FLOPs & Top-1 Acc. &\#Params &\#FLOPs & SS & MS &\#Params &\#FLOPs & AP$^{\text{box}}$ & AP$^{\text{mask}}$ \\
\midrule[0.5pt]
PoolFormer-S24\cite{metaformer} & pooling$\color{c1}{\bullet}$    &21M &3.5G &80.3 &51M &914G &42.3 &43.3 &41M &240G &41.5 &38.1 \\
\rowcolor{gray}
UniNeXt-T                    & pooling$\color{c1}{\bullet}$       &18M &3.3G &82.4 &47M &913G &45.7 &47.0 &38M &239G &45.2 &40.5 \\
\midrule[0.5pt]
ConvNeXt-T\cite{convnext} & 7$\times$7 conv$\color{c2}{\bullet}$  &29M &4.5G &82.1 &60M &939G &46.1 &46.7 &48M &262G &43.4 &39.5    \\
\rowcolor{gray}
UniNeXt-T                 & 7$\times$7 conv$\color{c2}{\bullet}$  &19M &3.4G &82.6 &48M &914G &46.7 &47.6 &38M &240G &45.8 &41.0   \\
\midrule[0.5pt] 
PVT-S\cite{PVT}    & spatial-reduction$\color{c3}{\bullet}$       &25M &3.8G &79.8 &54M &981G &43.7 &44.0 &44M &304G &40.4 &37.8  \\
\rowcolor{gray}
UniNeXt-T          & spatial-reduction$\color{c3}{\bullet}$       &29M &4.1G &83.1 &58M &994G &47.7 &48.3 &49M &317G &47.6 &42.6  \\
\midrule[0.5pt]
Swin-T\cite{Swin}  & shift window$\color{c3}{\bullet}$            &29M &4.5G &81.3 &60M &945G &44.5 &45.8 &48M &264G &43.7 &39.8  \\
\rowcolor{gray}
UniNeXt-T          & shift window$\color{c3}{\bullet}$            &24M &4.3G &83.5 &53M &938G &46.6 &47.9 &43M &263G &46.7 &41.9  \\
\midrule[0.5pt]
CSWin-T\cite{CSWin}  & cross-shaped window$\color{c3}{\bullet}$   &23M &4.3G &82.7 &60M &959G &49.3 &50.4 &42M &279G &46.7 &42.2  \\
\rowcolor{gray}
UniNeXt-T           & cross-shaped window$\color{c3}{\bullet}$    &24M &4.3G &83.5 &53M &961G &49.9 &50.5 &43M &282G &48.7 &43.6  \\
\midrule[0.5pt]
\rowcolor{gray}
UniNeXt-T              & local window(ours)$\color{c3}{\bullet}$  &24M &4.3G &83.6 &53M &942G &49.7 &50.6 &43M &266G &48.6 &43.4  \\
\bottomrule[0.15em]
 \end{tabular}}
\end{table*}

\begin{table*}[t]
\caption{\label{components}Ablation study on the effect of each component in our architecture. The base model in the first row is equipped with local window attention.}
\centering
\resizebox{1\linewidth}{!}{
\begin{tabular}{cccc|ccc|cccc|cccc}
\toprule[0.15em]
\multirow{2.2}{*}{HdC} &\multirow{2.2}{*}{EC} & \multirow{2.2}{*}{PC} &\multirow{2.2}{*}{Stem} &\multicolumn{3}{c}{ImageNet-1K} & \multicolumn{4}{c}{ADE20K} & \multicolumn{4}{c}{COCO} \\ 
& & & &\#Params &\#GFLOPs & Top-1 Acc. &\#Params &\#GFLOPs & SS & MS &\#Params &\#GFLOPs & AP$^{\text{box}}$ & AP$^{\text{mask}}$ \\
\midrule[0.5pt]

\xmark      & \xmark    & \xmark & \xmark &23.1M &3.9G & 81.3 & 52.9M  &933G     & 43.1 & 44.7 & 42.7M    &258G   & 42.6 & 39.0  \\
\cmark      & \xmark    & \xmark & \xmark &23.4M &4.0G & 82.5 & 53.2M  &935G     & 46.9 & 47.9 & 43.0M    &259G   & 46.5 & 42.0  \\
\cmark      & \cmark    & \xmark & \xmark &23.4M &4.0G & 82.9 & 53.2M  &935G     & 47.5 & 48.7 & 43.1M    &260G   & 47.1 & 42.5  \\
\cmark      & \cmark    & \cmark & \xmark &23.5M &4.0G & 83.1 & 53.3M  &931G     & 48.7 & 49.4 & 43.1M    &260G   & 47.7 & 43.0  \\
\cmark      & \cmark    & \cmark & \cmark &23.5M &4.3G & 83.6 & 53.3M  &942G     & 49.7 & 50.6 & 43.2M    &266G   & 48.6 & 43.4  \\
\bottomrule[0.15em]
 \end{tabular}}
\end{table*}

\subsection{Semantic Segmentation on ADE20K}
\noindent \textbf{Settings.} 
To demonstrate the effectiveness of our method on downstream tasks, we paired the widely used UperNet\cite{upernet} as a decoder and performed a fair comparison with another backbone on the ADE20k\cite{ADE20K} semantic segmentation dataset. All the models are trained for a total of 160k iterations with a batch size of 16. The AdamW\cite{AdamW} optimizer with a weight decay of 0.01 is used. The initial learning rate is set to 6e-5 and decay with a polynomial scheduler after the 1500-iterations warmup. Auxiliary losses are added to the output of stage 3 of the backbone with a factor of 0.4. We use the same data augmentation strategy as \cite{Swin,CSWin}.

\noindent \textbf{Results.} 
Table \ref{EXP_ade20k} shows the comparisons of different backbones on the ADE20K validation set. The single-scale (SS) and multi-scale (MS) mIoU are both reported. Our UniNeXt variants with local window attention outperform the state-of-the-art CSWin Transformer\cite{CSWin} by +0.4\%, +1.0\%, and +0.6\% SS mIoU, respectively. Besides, even with the plain local window attention, our UniNeXt-B achieves 51.4\%/52.2\% SS/MS mIoU. 
Finally, when UniNeXt is equipped with cross-shaped window attention, the base model achieves 51.6\%/52.5\% SS/MS mIoU. Furthermore, our results have surpassed many outstanding works\cite{vitcontroller, prseg, full, structtoken, ftn}. 
These results demonstrate the powerful context modeling capabilities of our UniNeXt.

\subsection{Object Detection and Instance Segmentation on COCO}
\noindent \textbf{Settings.} 
We evaluate UniNeXt performance on the COCO\cite{coco} dataset with the Mask R-CNN\cite{maskrcnn} architecture under 1x schedule (12 training epochs). We use AdamW\cite{AdamW} as the optimizer with a weight decay of 0.05 for all variants.
The learning rate is set to 0.0001 initially and decay at epoch 8 and 11 with a ratio of 0.1. The total batch size is 16. We follow Swin's\cite{Swin} data augmentation strategy.

\noindent \textbf{Results.} 
As shown in Table \ref{tab:det}, for object detection, with local window attention, our UniNeXt-T, UniNeXt-S, and UniNeXt-B achieve 48.6, 49.0, and 49.3 box mAP for object detection, surpassing the previous best CSWin Transformer\cite{CSWin} by +1.9, +1.1, and +0.6, respectively. In addition, our UniNeXt variants also achieve significant improvements in instance segmentation. For UniNeXt-T and UniNeXt-S, which are +1.2, +0.5 mask mAP higher than the previous best backbone.
When using cross-shaped window attention, our base model achieves 49.8 box mAP and 48.0 mask mAP.

\subsection{Ablation Study}
\subsubsection{Comparison with different token mixer}
Table \ref{ablation_attn_mode} shows the implementation of the different token mixers in our framework. From the experiment, we can find that for the token mixer with parameter-free (pooling), we are +2.1 Acc., +3.7 MS mIoU, +3.7 box mAP, and +2.4 mask mAP higher than the original implementation. 
For convolution (7$\times$7 conv), our implementation is +0.5 Acc., +0.9 MS mIoU, +2.4 box mAP, and +1.5 mask mAP higher than the ConvNeXt\cite{convnext}. 
For global attention (spatial-reduction), our implementation is +3.3 Acc., +4.3 MS mIoU, +7.2 box mAP, and +4.8 mask mAP higher than the PVT\cite{PVT}.
For local attention (shift window and cross-shaped window), our implementation is +2.2 Acc., +2.1 MS mIoU, +3.0 box mAP, and +2.1 mask mAP higher than the Swin\cite{Swin}, and +0.8 Acc., +0.1 MS mIoU, +2.0 box mAP, and +1.4 mask mAP higher than the CSWin\cite{CSWin}.
In addition, when using the plain local window attention, our UniNeXt-T achieves 83.6 Acc., 50.0 MS mIoU, 48.6 box mAP, and 43.4 mask mAP, surpassing the previous best performance.

It can be found that instantiating the token mixer with parameter-free, and attention in our framework leads to a stable performance improvement. When instantiating the token mixer with convolution, the whole framework becomes a fully convolutional architecture, and even in this case, our framework can bring some performance gains and outperform the previous best fully convolutional frameworks.

\subsubsection{Effect of components}
As shown in Table \ref{components}, we conducted ablation studies for each of the proposed convolution embedding modules. Each module not only improves the performance of the model on the classification, but also improves the performance on the downstream dense prediction task more significantly. A major performance boost is delivered by High-dimensional Convolution (HdC), which is +1.2 Acc., +3.8 SS mIoU, +3.2 MS mIoU, +3.9 box mAP and +3.0 mask mAP higher than the base architecture, with only +0.3M Params and +0.1G FLOPs. These lightweight modules can improve model performance significantly.

\subsubsection{Effect of window size}
Unlike image classification tasks, downstream dense prediction tasks often require a larger perceptive field.
We conducted an ablation study on the window size of the first three stages in local window attention, the last stage uses global attention. As shown in Table \ref{ablation_window_size}, as the window size increases (from 7\textbf{-}11), the performance of the model increases, and from 11\textbf{-}15, the performance of the model starts to decrease. So we choose window size\textbf{=}11 for dense prediction tasks.
\begin{table}[!h]
\caption{\label{ablation_window_size}Ablation study for different choices of window size. The pretrain model uses a window size of 7-7-7. The last stage uses global attention. The n-n-n represents the window size of stage1, stage2 and stage3.}
\centering
\resizebox{1\linewidth}{!}{
\begin{tabular}{c|ccc|ccc}
\toprule[0.15em]
\multirow{2.2}{*}{window size} & \multicolumn{3}{c}{ADE20K} & \multicolumn{3}{c}{COCO} \\
&\#FLOPs & SS & MS &\#FLOPs & AP$^{\text{box}}$ & AP$^{\text{mask}}$ \\
\midrule[0.5pt]

7-7-7   & 937G & 47.6 & 48.9       &263G     & 47.8 & 42.9  \\
9-9-9   & 939G & 47.3 & 48.8       &264G     & 48.3 & 43.3  \\
\rowcolor{gray}
11-11-11& 942G & 49.7 & 50.6       &266G     & 48.6 & 43.4  \\
13-13-13& 943G & 48.9 & 49.7       &268G     & 48.4 & 43.5  \\
15-15-15& 947G & 47.8 & 49.6       &272G     & 48.4 & 43.6   \\
\bottomrule[0.15em]
 \end{tabular}}
\end{table}

\subsubsection{Efficient evaluation for dwconv}
\begin{figure}[!h]
\centering
\includegraphics[width=1\linewidth]{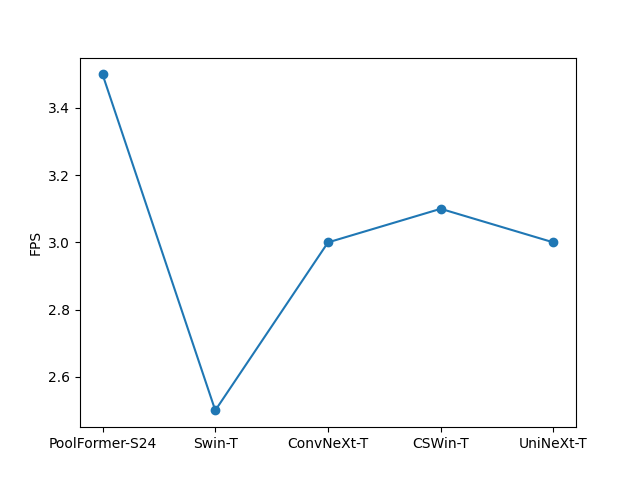} 
\caption{\label{fps} The real-time speed comparison.}
\end{figure}

As shown in Table \ref{fps}, We conducted a real-time speed comparison of NVIDIA's 1050-Ti GPU and found that the difference in speed between UniNeXt and other Transformers is negligible. Additionally, we removed the modules that utilized dwconv in UniNext (HdC, EC, and PC) resulting in an increase in FPS from 3.0 to 3.3. However, we believe that this difference is acceptable.

\subsubsection{Evaluation for position embedding}
The results are displayed in Table \ref{pe}, which reveals that incorporating positional embedding leads to a decline in performance. As the proposed module already has the capability to incorporate implicit positional embedding, there is no need for additional explicit positional embedding.

\begin{table}[!h]
\caption{\label{pe} Evaluation for position embedding. }
\centering
\resizebox{0.6\linewidth}{!}{
\begin{tabular}{ccc|c}
\toprule[0.15em]
UniNeXt-T & APE & RPE &Top-1 Acc.  \\
\midrule[0.6pt]
\cmark &\xmark&\xmark& 83.6 \\
\cmark &\cmark&\xmark& 83.4 \\
\cmark &\xmark&\cmark& 83.5 \\
\cmark &\cmark&\cmark& 83.6 \\
\bottomrule[0.15em]
 \end{tabular}}
\end{table}

\subsubsection{Effective Receptive Field}
\begin{figure}[!h]
\centering
\includegraphics[width=1\linewidth]{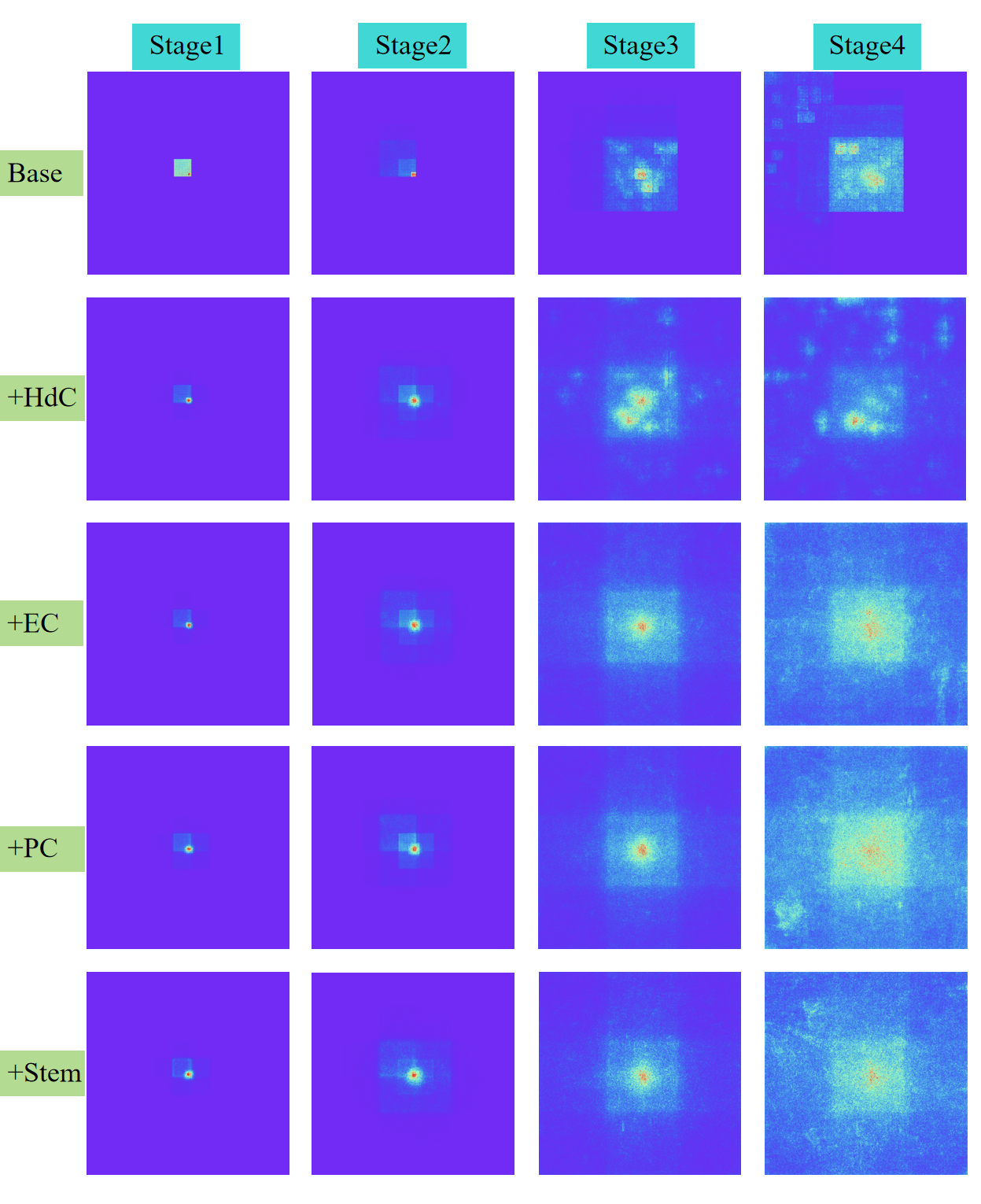} 
\caption{Effective Receptive Field (ERF) on ADE20K (average over 100 images). ERFs of the four stages are visualized. The ``Base'' model represents the UniNeXt without any convolution embedding and using the local window attention of 11$\times$11 window size. The convolution embedding modules are added one by one, ``+HdC'' means the model is ``Base+HdC'', ``+EC'' means the model is ``Base+Hdc+EC'', and so on. Best viewed with zoom in.}
\label{erf}
\end{figure}
For dense prediction tasks (e.g., segmentation and detection), fusing more context information with a larger receptive field has been a central issue.
We use the effective receptive field (ERF)\cite{erf} as a visualization tool to explain the effectiveness of the proposed convolution embedding module. In Figure \ref{erf}, we visualize the ERFs of the four encoder stages. All models use local window attention. 
We can obtain the following observations:
\begin{itemize}
\item Even using local window attention with a window size of 11$\times$11, the proposed convolution embedding module (i.e., each convolution embedding module uses a 3$\times$3 convolutional kernel) can bring a stable effective receptive field enhancement.
\item The addition of the convolution embedding enables more focused local attention, as shown in the stage3 and stage4 columns, where the receptive fields converge on the center from top to bottom.
\end{itemize}

\section{Conclusion}
In this paper, unlike designing an advanced spatial token mixer, we find that the design of the overall architecture is more important. We extends the double-jump connection architecture to the multi-jump connection architecture, formed a unified architecture for various spatial token mixers, called UniNeXt. We found some important conclusions through extensive experiments and effectively proved the effectiveness of the proposed framework. Finally, we expect the community to focus on macroscopic architecture design rather than just spatial token mixers.

\section{Acknowledgment}
This work was supported by Damo Academy through Damo Academy Research Intern Program.

\balance

\bibliographystyle{ACM-Reference-Format}
\bibliography{sample-base}

\appendix

\end{document}